# Automatic Keyword Extraction from Spoken Text.
## A Comparison of two Lexical Resources: the EDR and WordNet.

**Lonneke van der Plas[1], Vincenzo Pallotta[2], Martin Rajman[2], Hatem Ghorbel[2]**

[1]Rijksuniversiteit Groningen
Informatiekunde
vdplas@let.rug.nl

[2]Swiss Federal Institute of Technology - Lausanne
Faculty of Information and Computer Science
{Vincenzo.Pallotta, Martin.Rajman, Hatem.Ghorbel}@epfl.ch

**Abstract**

Lexical resources such as WordNet and the EDR electronic dictionary (EDR) have been used in several NLP tasks. Probably partly due to the fact that the EDR is not freely available, WordNet has been used far more often than the EDR. We have used both resources on the same task in order to make a comparison possible. The task is automatic assignment of keywords to multi-party dialogue episodes (i.e. thematically coherent stretches of spoken text). We show that the use of lexical resources in such a task results in slightly higher performances than the use of a purely statistically based method.

## Introduction

Lexical resources, such as WordNet (Fellbaum, 1998) and the EDR electronic dictionary (Miike et al., 1990), are used in many NLP tasks, both in multilingual environments where one needs translations of words, and in monolingual environments. For instance, WordNet has been used for text summarization (Barzilay and Elhadad, 1997), and EDR for semantic word organization (Murata et al. 2000). In our experiments, we have used both WordNet and EDR for the same task: automatic extraction of important keywords, i.e. the words that reflect best the subject of the processed documents.

For this task, we used the information about IS-A relations that can be found in both lexical resources[1]. An example of an IS-A relation is the relation between 'cat' and 'animal (i.e. IS-A(cat, animal)). These IS-A relations can be used in computing the semantic similarity between words. Several measures have been developed to assign values of semantic similarity. A good overview of these measures for WordNet, together with their application to Word Sense Disambiguation can be found in (Budanitsky and Hirst, 2001) and (Patwardhan et al. 2003). For the particular task discussed in this paper we have chosen the Leacock-Chorodow similarity measure (Leacock and Chodorow, 1998).

The main goal of our experiments was to use the keyword extraction task to perform a comparative evaluation of the performance achieved with different lexical resources. In addition, we have compared the method using the lexical resources with a very simple baseline approach that only takes into account word frequencies (more precisely the RFR score as described later on).

In the rest of the paper, we first shortly describe the lexical resources and the textual data we have used for our experiments. Then, we describe the evaluated keyword extraction methods and the comparative evaluation set up. Finally, we provide and discuss the obtained results and propose a conclusion and some ideas for future work.

---

[1] The IS-A relation corresponds to the *hyperonymy* relation in WordNet.

## Available lexical resources

**WordNet.** WordNet (Miller, 1998) is a freely available electronic dictionary developed by the Cognitive Science Laboratory at Princeton University. Its design is inspired by current psycholinguistic theories of human lexical memory. Words are organized into synonyms sets (i.e. *synsets*) each representing one underlying lexical concept. For example: the set of lexical items {car, automobile auto, machine, motorcar} constitutes one synset representing the concept corresponding to the *gloss* definition "4-wheeled motor vehicle; usually propelled by an internal combustion engine". Different semantic relations link synsets together into several hierarchies (e.g. IS-A and PART-OF relations). WordNet 2.0 contains 93.196 synsets (Yu et al., 2003).

**EDR.** The EDR electronic dictionary is a bilingual large-scale lexical database developed by the Electronic Dictionary Research Institute in Japan (Miike et al., 1990). The EDR dictionary has several components: an English and a Japanese word dictionary, an English and a Japanese co-occurrence dictionary, two bilingual dictionaries (JAP->EN and EN->JAP) and a *Concept* dictionary. It is this last dictionary that is most interesting for our task. Each word that can be found in the EDR dictionaries is linked to one or more concepts (polysemous words). A gloss definition is attached to most of the concepts. For example, the word forms: *auto, car, autocar, motorcar, motocar and wheel* all are linked to the concept *automobile* for which the definition is: "four-wheeled vehicle called an automobile". The Concept Dictionary contains the Concept Classification Dictionary. This Classification Dictionary contains the classification of concepts that are in an IS-A relation. EDR contains 488.732 concepts.

## The Meeting Dialogues Data

The corpus we have used for our experiments is the ICSI collection of meeting dialogues transcriptions (Janin et al., 2003). In these multi-party dialogues, an average of 6 participants are talking about topics related to speech recognition and audio equipment. 25 of these transcriptions have been segmented at Columbia University (Galley et al., 2003). We selected dialogues for which the most fine-grained segmentation was available. Finally, we used 6 dialogues, containing on average 9 segments and 38.000 words.

## Methods and Algorithms

For our keyword extraction task, we have limited ourselves to simple nouns. Reason for this is that nouns are very informative and the IS-A hierarchy for nouns in conceptual dictionaries is better structured. For each segment the text is first tagged with the *TreeTagger* (Schmid, 1994), a probabilistic decision trees based Part-of-Speech tagger. We then select all simple nouns according to the assigned PoS tags[2].

As mentioned above, a first intuitive criterion for finding important keywords in a text is frequency. To this extent, we have used the *relative frequency ratio* (RFR) (Damerau, 1993). For a given word, RFR is the ratio of the frequency of the word in the document to the frequency of the word in a general corpus (BNC corpus). The idea is that words that occur relatively frequently in a certain text compared to how frequently they occur in a general text are more likely to be good keywords. We have used the relative frequency of a word as an initial filter. Words that have an RFR below a certain threshold (10 in our case) are filtered out.

The second criterion we have used to find important keywords relies on a measure of semantic similarity derived from the available lexical resources. Each word that has passed the RFR filter is first associated with a concept. As, both in EDR and WordNet, words can be associated with multiple concepts (polysemy), we have adopted a simple word sense disambiguation strategy[3]: we have selected the most common sense for each word and used the corresponding concept for further processing. For WordNet we have selected the first synset associated with the word under consideration. For EDR, since a frequency is attached to each concept associated with a particular word, we selected the concept with the highest frequency.

To measure the semantic similarity between two concepts, we have used the Leacock-Chorodow concept similarity measure (Leacock and Chodorow, 1998). This measure is based on the length of the paths between pairs of concepts in a ISA-hierarchy, scaled according to the depth of the IS-A hierarchy as a whole. More precisely, the measure of semantic similarity is defined thus as follows:

$$SS(c_1,c_2) = \max[-\ln(SP(c_1,c_2)/2*D)]$$

where $SP(c_1,c_2)$ is the *shortest path* between the two concepts $c_1$ and $c_2$ (i.e. the path connecting them that has the least number of nodes in the IS-A hierarchy) and, $D$ is a constant estimated as the maximal *depth* of the IS-A hierarchy. For both our lexical resources, $D$ has been set to 16.

---

[2] The impact of the accuracy of the tagger, quite strongly affected by the fact that spoken data was used, on the overall keyword extraction performance has not been assessed yet.
[3] We are aware that Word Sense Disambiguation plays a major role in the overall performance of the system, which has not been yet adequately assessed.

The semantic similarity between a concept and itself (self-similarity) cannot be accounted for with the above formula (as the log is not defined for SP=0). Therefore, specific values need to be defined for the self-similarities. We have chosen the simplest solution, where all self-similarities are equal to a constant $SS_0$, i.e. for any concept $c$, $SS(c,c)=SS_0$. To guarantee that $SS()$ remains a similarity, $SS_0$ has to verify that, for any pair of concepts $\{c_1,c_2\}$, $SS(c_1,c_2) \leq SS_0$. As, in our case, the maximal value for $SS(c_1,c_2)$ is 3.47 (obtained for $SP(c_1,c_2)=2$, the minimal path length between two leaves of the ISA-hierarchy), we have chosen $SS_0 = 5$.

Once a semantic similarity was defined for the concepts, we used the following procedure to extract the keyword candidates:
(1) first we performed a (hierarchical) single-link clustering (Theodoridis and Koutroumbas, 2003) to find clusters of concepts related to each other with decreasing values of $SS$ (up to a predefined lower threshold). The used clustering algorithm is briefly described below.
(2) then, for each of the produced clusters (resp. each of the concepts within each of the clusters), we computed a *score*, hereafter referred to as the *cluster level* score (resp. the *concept level* score). Both scores (formally defined below) were used to derived a global ranking on the concepts by simply first taking into account the cluster level score, and then (i.e. for all the concepts in the same cluster) taking into account the concept level score.
(3) finally, we used the resulting concept ranking to produce a list of "best" concept candidates, and, for each of these concept candidates, we produced the associated word(s) as keyword candidates. Notice that we tried several methods to define the number of candidate keywords to be selected. This point is discussed below in the "evaluation framework" section.

*The cluster level score*: this score is defined as the sum of two components:

- the *connectivity* ($Cn$), corresponding to a global measure of the semantic relatedness between all concepts in the cluster, and defined, for any cluster $CL$, as the sum of the semantic similarities between all pairs of concepts in the cluster:

$$Cn(CL) = \sum_{c_i,c_j \in CL, i<j} SS(c_i,c_j)$$

- the *reiteration (R)*, corresponding to a global measure of the importance of the concepts in the cluster, and defined, for any cluster CL, as:

$$R(CL) = SS_0 \sum_{c \in CL}(n(c)-1)$$

where $n(c)$ is the number of occurrences of the word(s) associated with the concept $c$ in the text.

*The concept level score*: this score is defined in a similar way as the sum of two components: the *concept level connectivity*, defined, for any concept $c_i$ in a cluster $CL$, as the sum of the semantic similarities between $c_i$ and all the other concepts in $CL$; and the *concept level reiteration*, defined for any concept $c$ as the product $SS_0(n(c)-1)$. Formally, the concept level score $CS_{CL}(c)$ of a concept $c$ in a cluster $CL$ is therefore defined as:

$$CS_{CL}(c) = (\sum_{c_i \in CL, c \neq c_i} SS(c,c_i)) + SS_0(n(c)-1)$$

Notice also that, in addition to the above scores, we used the RFR values to make the selection between candidate keywords of the same final rank: the keyword with the highest RFR was systematically preferred.

## Comparative Evaluation

### Evaluation framework and results

For the evaluation, we used manually selected keywords as a gold standard. Four annotators (including the first author but for less than 20% of the data) were asked to select keywords (only nouns, no compounds) for each of the dialogue segments. To make this task easier, the lists of all nouns found in each of the segments were automatically generated and the annotators were asked to prioritarily choose keywords from these list. However, they also had the possibility to add missing keywords, if really necessary. Notice that the keywords that were actually added were rare and often corresponded to words other than nouns. Therefore, the added keywords were not taken into account for the evaluation.

To measure the impact of the constraint of choosing only nouns, we also carried out a complementary experiment (with only three dialogues), where the annotators were allowed to choose from keywords lists containing both simple nouns and multi-word expressions[4]. As it might have been expected, this experiment clearly showed that annotators have indeed the tendency to choose multi-word expressions (i.e. 26% of the chosen keywords).

No strong constraint was imposed on the number of keywords to associate with a segment. The annotators were free to choose up to 10 keywords, as we felt that this corresponds to what people typically do. The number of keywords associated with each of the segments varied between 3 and 10, with an average of 6.2.

As a general remark on the evaluation setup, the annotators found the task very difficult. This was due to the nature of the data (transcribed spoken data) and the topic of the dialogues (i.e. speech recognition). It was considered as very tiring to read transcriptions where people often did not finish their sentences and used technical jargon. Moreover, it was difficult to follow who is talking to whom, and to deal with some segments, which were quite large.

To produce the evaluation results, we used two types of evaluation metrics:

---
[4] noun-noun, adj-noun, adj-noun-noun, adj-adj-noun combinations.

- *average k-accuracy*: the average number of correct keyword propositions in the k-best keyword candidates produced for each segment by the system, with k being the number of keywords assign to the segment by the annotator. Because the number of keywords was not imposed, k varied between 3 and 10.

- *average precision, recall* and *F-measure*: in this case no a-priori knowledge about the number of keywords assigned by the annotators was assumed; we therefore let the system systematically produce $k$=2, 5, and 10 candidate keywords for each segment, and, for each of the $k$ values, we used the gold standard to compute standard precision, recall and F scores.

In the result table 1 below, the average $k$-accuracy scores are associated with the value "variable" in the column "#keywords" (number of keywords), while the average Precision, Recall, and *F*-measure (in the form *P/R/F*) are associated with the corresponding value of $k$. The three columns containing the average scores correspond to the results obtained when using the semantic similarities computed with the IS-A hierarchy from EDR (EDR) or from WordNet (WN), with different values (1.85, 1.67, or 1.51) for the lower semantic similarity threshold used for the clustering.

Table 1. Keyword extraction performance with EDR and WordNet

| #keywords | EDR1.85 | EDR1.67 | EDR1.51 |
|---|---|---|---|
| 2 | .48/.18/.24 | .48/.16/.24 | .53/.19/.26 |
| 5 | .40/.32/.35 | .39/.31/.34 | .42/.34/.37 |
| 10 | .38/.43/.39 | .38/.45/.40 | .36/.48/.40 |
| variable | .39 | .36 | .38 |
| **#keywords** | **WN1.85** | **WN1.67** | **WN1.51** |
| 2 | .60/.21/**.31** | .43/.15/.22 | .42/.15/.22 |
| 5 | .49/.35/**.41** | .34/.28/.30 | .33/.26/.30 |
| 10 | .43/.44/**.42** | .31/.40/.34 | .30/.40/.34 |
| variable | **.47** | .33 | .32 |

The results obtained with the baseline approach simply using the RFR scores are also provided for comparison in table 2.

Table 2. Keyword extraction performance with RFR

| #keywords | RFR |
|---|---|
| 2 | .30/.12/.17 |
| 5 | .32/.30/.30 |
| 10 | .30/.51/.38 |
| Variable | .33 |

### Discussion of the results

The first interesting result that can be clearly derived from table 1 is that WordNet produces the best performances overall and that all its best scores were obtained for the highest value (1.85) for the semantic similarity threshold. However, at the other semantic similarity thresholds (1.67, 1.51), EDR outperforms WordNet.

Then, it can be noticed that, in general, WordNet shows a higher precision, while EDR has a higher recall.

Another interesting observation is that, with WordNet, performance decreases when the threshold for semantic similarity decreases, while EDR is less affected by lowering the threshold for similarity and scores best for the lowest semantic similarity threshold (1.51). As EDR is large-scale resource, substantially larger than WordNet that contains far less relations, a possible interpretation is that the optimal threshold for a resource depends on its depth and branching factor. In any case, this result stresses the fact that one should be very cautious in setting the threshold parameter for a given lexical resource.

As far as the number $k$ of selected candidate keywords is concerned, it can be observed that the F-measures culminate for $k$=10 for both WordNet and EDR. However, 10 keywords seem too much to capture the topic of a segment. Since participants selected an average of 6.2 keywords a threshold of 5 candidate keywords seems more suitable. Under this assumption, WordNet (at 1.85) performs slightly better than EDR (1.51).

As far as $k$-accuracy is concerned, it seems that WordNet performance decreases at lower semantic similarity thresholds, while EDR performances tend to be stable.

Finally, both lexical resources outperform the purely frequency based method (RFR). This seems to indicate that RFR might not be the most suitable method for selecting keywords. However, other statistical methods such as log-likelihood (Dunning, 1993) might result in better scores.

## Conclusion and future work

In this contribution, we have shown that lexical resources can provide valuable performance increase for quite useful NLP tasks such as keyword extraction from spoken dialogues. Our work also showed, however, that the used methods and the parameters need to be adapted to the nature of the available resources. Finally, no really significant difference in performance was observed between the two resources (EDR and WordNet), when the parameters (i.e. the SS thresholds) are optimally chosen.

It would be interesting to run the same experiments with different data. In our experiments, we used a corpus of spoken multi-party meeting dialogues that was difficult to process and difficult to read for our annotators. The results might be different in the case of written text corpora and hence more interesting for the comparison of resources.

The method used for scoring could also be changed. Currently, we are considering two components for computing the scores: connectivity and reiteration. Reiteration might also be taken into account in another way by explicitly adding, for each reiterated word, its corresponding concept in the clusters, therefore leading to concept reiteration, instead of word reiteration. In this way, both the connectivity and the reiteration would relate to concepts and might therefore be taken into account in a uniform way, which might correspond to a theoretically sounder approach.

## Acknowledgements

The work presented here is part of the Swiss NCCR on "Interactive Multimodal Information Management" (IM2, http://www.im2.ch), funded by the Swiss National Science Foundation. The work pertains specifically to the IM2.MDM[5] IP, on "Multimodal Dialog Management".

---

[5] http://www.issco.unige.ch/projects/im2/mdm.